\documentclass[journal]{IEEEtran}

\usepackage{times}
\usepackage{epsfig}
\usepackage{graphicx}
\usepackage{amsmath}
\usepackage{amssymb}
\usepackage{color}
\usepackage{stmaryrd}
\usepackage{cite}
\usepackage{booktabs}
\usepackage{enumerate}
\usepackage{sidecap}
\usepackage{multirow,bigdelim}
\usepackage[table]{xcolor}

\usepackage[breaklinks=true,bookmarks=false]{hyperref}

\newcommand{\etal}{{\it et al.}}

\ifCLASSINFOpdf
\else
\fi

\begin{document}

\title{Regularization of Deep Neural Networks with Spectral Dropout}

\author{Salman~H~Khan,~\IEEEmembership{Member,~IEEE,}
        Munawar~Hayat,~
        and~Fatih~Porikli,~\IEEEmembership{Fellow,~IEEE}
\thanks{S. H. Khan is with the Data61, Commonwealth Scientific and Industrial Research Organisation (CSIRO),  Canberra ACT 2601 and Australian National University, Canberra ACT 0200, e-mail: salman.khan@csiro.au}
\thanks{M. Hayat is with the University of Canberra, Bruce ACT 2617.}
\thanks{F. Porikli is with the Australian National University, Canberra ACT 0200.}
}

\markboth{}
{Shell \MakeLowercase{\textit{et al.}}: Bare Demo of IEEEtran.cls for IEEE Journals}

\maketitle

\begin{abstract}
The big breakthrough on the ImageNet challenge in 2012 was partially due to the `dropout' technique used to avoid overfitting. Here, we introduce a new approach called `Spectral Dropout' to improve the generalization ability of deep neural networks. We cast the proposed approach in the form of regular Convolutional Neural Network (CNN) weight layers using a decorrelation transform with fixed basis functions. Our spectral dropout method prevents overfitting by eliminating weak and `noisy' Fourier domain coefficients of the neural network activations, leading to remarkably better results than the current regularization methods. Furthermore, the proposed is very efficient due to the fixed basis functions used for spectral transformation. In particular, compared to Dropout and Drop-Connect, our method significantly speeds up the network convergence rate during the training process (roughly $\times2$), with considerably higher neuron pruning rates (an increase of $\sim 30\%$). We demonstrate that the spectral dropout can also be used in conjunction with other regularization approaches resulting in additional performance gains. 

\end{abstract}


\IEEEpeerreviewmaketitle

\section{Introduction}
\IEEEPARstart{D}{eep} neural networks with a huge number of learnable parameters are prone to overfitting problems when trained on a relatively small training set. This leads to poor performance on the held-out test data because the learned weights are tailored only for the training set, and they lack the generalization ability to unseen data. It has been observed that the overfitting problem is caused due to complex co-adaptations of the neurons which make the neural network dependent on their joint response instead of favoring each neuron to learn a useful feature representation \cite{hinton2012improving}. A number of simple, yet powerful methods have been designed over the recent years to prevent overfitting during the network training. These methods include data augmentation \cite{simard2003best}, $\ell 1$ and $\ell2$ regularization \cite{orr2003neural}, elastic net regularization \cite{zou2005regularization}, weight decay \cite{moody1995simple}, early stopping \cite{morgan1989generalization}, max-norm constraints, and random dropout \cite{srivastava2014dropout}. 

\begin{figure}
\centering
\includegraphics[width=0.9\columnwidth]{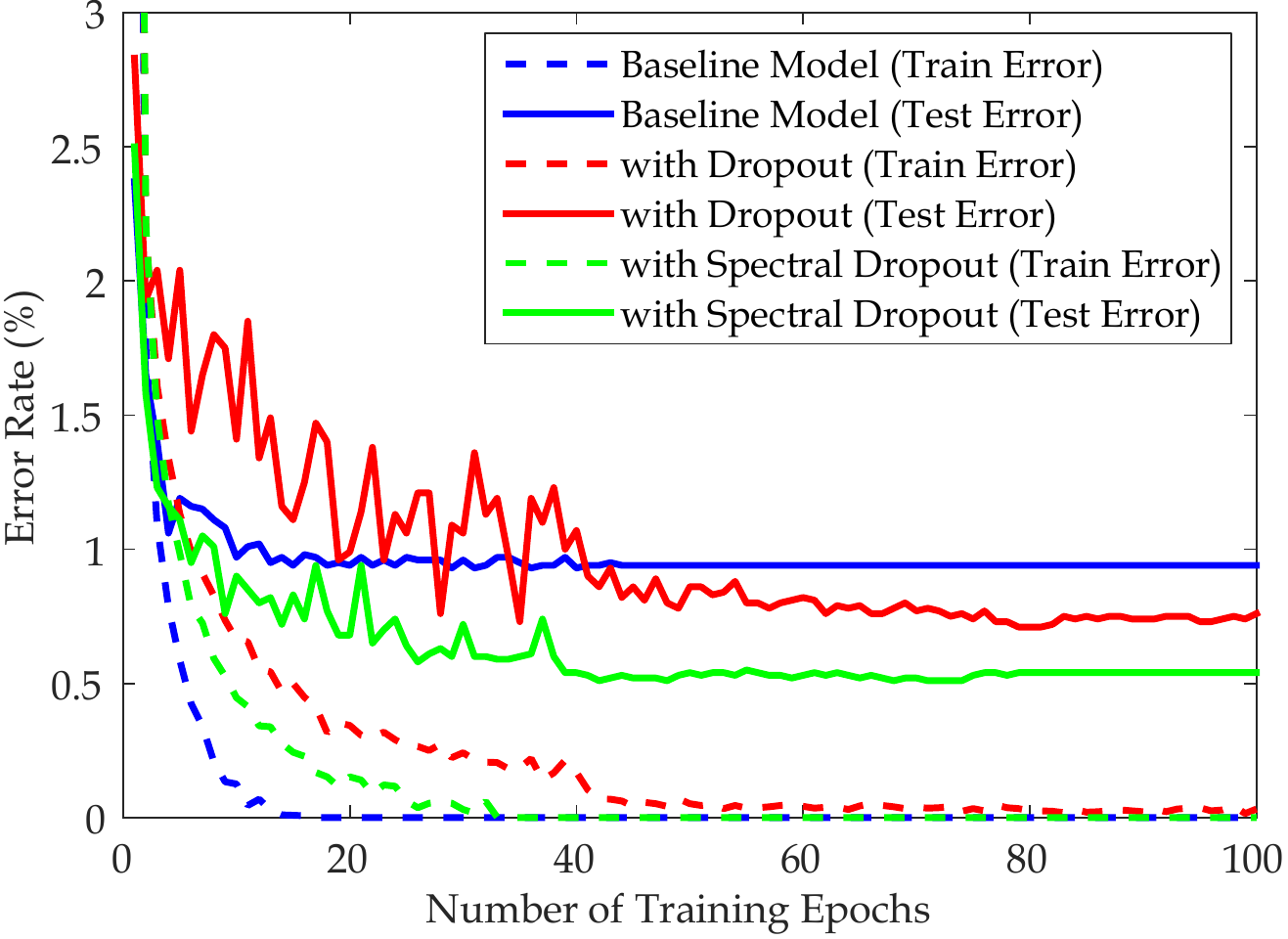}
\caption{\textbf{Spectral Dropout} improves the network generalization ability and yields better performance on unseen data. Further, the convergence speed is increased approximately by a factor of two compared to random dropout. The error plots are shown for training on the MNIST dataset using LeNet architecture. }
\label{fig:mnist_err}
\end{figure}

A general theme to enhance the generalization ability of neural networks has been to impose stochastic behavior in the network's forward data propagation phase. Examples of such schemes include Dropout, which randomly shuts down neurons \cite{srivastava2014dropout}, Drop-Connect, which randomly deactivates connections between neurons \cite{wan2013regularization}, spatial shuffling, which randomly performs block-wise image reorganization \cite{hayat2015spatial}, and fractional max-pooling, which randomly changes the pooling region \cite{graham2014fractional}. Generally, these approaches also perform a marginalization step during the prediction phase to compute the expected output. Data augmentation (e.g., with color jittering \cite{krizhevsky2012imagenet}), stochastic pooling \cite{zeiler2013stochastic} and model averaging \cite{he2015delving} to create an ensemble effect can also be interpreted in a similar way. All these regularization approaches force the network to learn generic feature detectors, instead of merely memorizing the training samples.  

In this paper, we propose a different approach for network regularization that does not adopt a fully randomized procedure, yet achieves improved generalization by dropping the noisy spectral components (see Fig.~\ref{fig:mnist_err}). To avoid co-adaptations of the feature detectors and to identify noisy spectral components, we propose to use a decorrelation transform, such as the discrete cosine transform (DCT). In contrast to the Dropout approach that randomly shuts down feature detectors during the training phase, our approach drops out the less significant spectral components to preserve the discriminative ability of network activations. This enables the network to become invariant to the `noisy' spectral components by randomly selecting only the most important basis vectors for signal reconstruction during the spectral dropout regularization process. Furthermore, while Dropout slows down the network convergence speed roughly by a factor of two or more \cite{krizhevsky2012imagenet}, our spectral dropout approach does not impede the network convergence rate and therefore provides gain in both the network performance and training efficiency.  

Spectral dropout complements other regularization techniques (e.g., random dropout, $\ell1/\ell2$ regularization) and can easily work in conjunction with them to achieve a better performance and superior generalization capability. We demonstrate that the proposed approach generates a very compact and sparse intermediate feature representation that can significantly reduce the storage requirements for applications which perform retrieval, comparisons or matching in the feature space. Our experiments extensively test the proposed approach with different popular network architectures such as the LeNet, Network in Network (NiN) and the recent Residual Network (ResNet). The experimental analysis shows significant improvements in classification performance on the MNIST, CIFAR and SVHN datasets achieving low error rates of $0.38\%$, $5.76\%$ and $2.12\%$ respectively.

The main highlights of this paper include:
\begin{itemize}
\setlength{\itemsep}{0pt}
\setlength{\parskip}{0pt}
\setlength{\parsep}{0pt}   
\item A novel approach to reduce network overfitting using spectral dropout.
\item Despite of much higher neuron pruning rates, our approach achieves better convergence performance compared to Dropout and Drop-Connect. 
\item Spectral dropout can be used to obtain uncertainty estimates during the test phase which can signify the confidence of network predictions. 
\item Our experiments show a consistent and remarkable performance boost on a diverse set of networks. 
\end{itemize}

The rest of the paper is organized as follows. A brief overview of the closely related approaches is given in Sec.~\ref{sec:Background}. The proposed approach is described in Sec.~\ref{sec:SpectralDropout} followed by a detailed literature review in Sec.~\ref{sec:RelatedWork}. The specific implementation details are given in Sec.~\ref{sec:cnn_arch} and our experimental results are given in Sec.~\ref{sec:exp_analysis}. We provide a thorough analysis on various aspects of our approach in Sec.~\ref{sec:analysis}.

\section{Related Work}\label{sec:RelatedWork}
\textbf{Network Regularization} has been an active research area recently. We review in detail two very popular approaches namely Dropout \cite{srivastava2014dropout} and Drop-Connect \cite{wan2013regularization} in Sec.~\ref{sec:Background}. Batch normalization is another related approach that rather indirectly improves network generalization by reducing internal covariance shift of the feature representations \cite{ioffe2015batch}. In this aspect, batch normalization is close to data decorrelation and whitening based approaches which have been traditionally used for automatic feature learning \cite{krizhevsky2009learning}. Other remarkable regularization techniques include normalization \cite{orr2003neural}, weight decay \cite{moody1995simple}, model averaging \cite{he2015delving}, early stopping \cite{morgan1989generalization}, Gaussian dropout \cite{wang2013fast}, and sparse constraints\cite{porikli2016less}. Huang \etal recently proposed a stochastic approach to vary network depth during training, which performs surprisingly well in practice \cite{huang2016deep}. Different to above mentioned approaches, our work proposes a new regularization framework based on spectral representations. 

\textbf{Spectral Representations} have been proposed in the literature  to achieve computational advantages in the learning and inference of deep neural networks \cite{ben1999fast}. Mathieu \etal \cite{mathieu2013fast} used the Fourier transform to speed-up the expensive convolution operation in the spatial domain. Similar approaches have been reported in \cite{lavin2015fast,vasilache2014fast} to enhance the network efficiency. In addition to enabling fast computations, frequency domain representations have been used to reduce the storage and memory requirements in deep networks \cite{sindhwani2015structured}. This is made possible due to the fact that network parameters can be compactly represented in the spectral domain removing any redundancy \cite{chen2015compressing,cheng2015exploration}. 
 More recently, Rippel \etal \cite{rippel2015spectral} proposed spectral representations for activation pooling and parametrization to achieve dimensionality reduction and better network optimization. 
In contrast to these works, we use spectral representations to enhance the generalization ability of deep networks without compromising on the convergence performance. Note that \cite{rippel2015spectral} also achieves faster convergence rates by optimizing filters in the spectral domain. However, they need to switch many times back and forth between spatial and spectral domains to apply spatial-domain non-linearities. This repeated forward and inverse transformations are very expensive and therefore their faster convergence rate does not imply overall reduction in learning time. In contrast, our approach enhances convergence speed while working with spatial domain filters and avoids such additional overhead. 

\section{Background}\label{sec:Background}

One of the most popular approaches for neural network regularization is the Dropout technique \cite{srivastava2014dropout}. During network training, each neuron is activated with a fixed probability (usually 0.5 or set using a validation set). This random sampling of a sub-network with-in the full-scale network introduces an ensemble effect during the testing phase, where the full network is used to perform prediction. Another similar approach is the Drop-Connect \cite{wan2013regularization}, which randomly deactivates the network weights (or connections between neurons) instead of randomly reducing the neuron activations to zero. In contrast to random Dropout and Drop-Connect, our approach regularizes the network output by discarding noisy spectral components during the train and test phase.


Let us consider a CNN that is composed of $L$ weight layers, indexed by $l \in \{1 \ldots L\}$. Since Dropout and Drop-Connect have predominantly been applied to Fully Connected (FC) layers in the literature, we consider the simpler case of FC layers first. Given output activations $\mathbf{a}_{l-1}$ from the previous layer, a FC layer performs an affine transformation followed by a element-wise non-linearity, as follows:
\begin{align}
\mathbf{a}_{l} = f(\mathbf{W} * \mathbf{a}_{l-1} + \mathbf{b}_{l}).
\end{align}
Here, $\mathbf{a}_{l-1} \in \mathbb{R}^{n}$ and $\mathbf{b} \in \mathbb{R}^{m}$ denote the activations and biases respectively, $\mathbf{W} \in \mathbb{R}^{m\times n}$ is the weight matrix and $f(\cdot)$ is the  Rectified  Linear  Unit  (ReLU)  activation  function. 

\begin{figure*}
\centering
\includegraphics[clip=true, trim= 0mm 10cm 0mm 3cm, width = \textwidth]{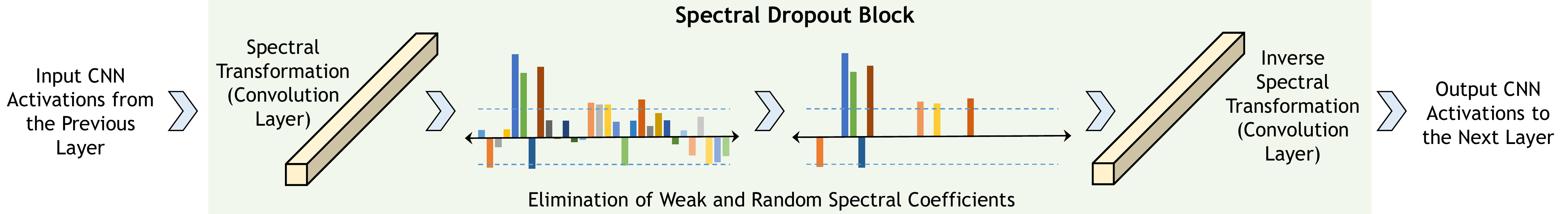}
\caption{Spectral Dropout Block with in a CNN enhances network's generalization ability by reducing spectral coefficients.}
\label{fig:SDblock}
\end{figure*}

\paragraph{Dropout:}
The random dropout layer generates a mask $\mathbf{m} \in \mathbb{B}^{m}$, where each element $m_i$ is independently sampled from a Bernoulli distribution with a probability `$p$' of being on: 
\begin{align}
m_i \sim Bernoulli(p), \qquad  m_i \in \mathbf{m}.
\end{align}
This mask is used to modify the output activations $\mathbf{a}_l$:
\begin{align}
\mathbf{a}_{l} = \mathbf{m} \circ f(\mathbf{W} * \mathbf{a}_{l-1} + \mathbf{b}_{l}),
\end{align}
where, `$\circ$' denotes the Hadamard product.

\paragraph{Drop-Connect:}
Similar to Dropout, Drop-Connect performs masking out operation on the weight matrix instead of the output activations, therefore: 
\begin{align}
\mathbf{a}_{l} & = f((\mathbf{M} \circ \mathbf{W}) * \mathbf{a}_{l-1} + \mathbf{b}_{l}), \\
M_{i,j} & \sim Bernoulli(p), \qquad  M_{i,j} \in \mathbf{M}.
\end{align}
Next, we describe the proposed spectral dropout approach.

\section{Spectral Dropout}\label{sec:SpectralDropout}

The spectral dropout approach ignores the noisy and weak frequency domain coefficients corresponding to activations from a CNN layer (Fig. \ref{fig:SDblock}). Consequently, three key benefits are achieved: \emph{First}, it provides an effective way to perform regularization by discouraging co-adaptations of feature detectors. \emph{Second}, the network convergence rate during training is increased, roughly by a factor of two, compared to the regular dropout. This is because a significant portion of strong frequency coefficients are retained for signal reconstruction. \emph{Third}, frequency dropout can be applied after any neural network layer to enhance the network generalization. This in contrast to Dropout and Drop-Connect, which are usually applied to the final FC layers (see Sec.~\ref{sec:Background}).

Let us consider $\mathbf{A}_{l-1} \in \mathbb{R}^{h\times w \times n}$ to be a tensor representing output CNN activations from the previous layer. We can represent convolutional filters as $\mathbf{F}_{l} \in \mathbb{R}^{h' \times w' \times n \times m}$ which operate on $\mathbf{A}_{l-1}$ to give output activations $\mathbf{A}'_{l}\in \mathbb{R}^{h''\times w'' \times m}$, as follows:
\begin{align}
\mathbf{A}'_l = f(\mathbf{F}_{l} \otimes \mathbf{A}_{l-1} + \mathbf{b}_{l}).
\end{align}
The above expression denotes the normal operation of a convolutional or a FC layer. For the spectral dropout, we need to perform frequency domain transformation followed by the truncation of noisy coefficients and inverse transformation to reconstruct the original CNN activations from the last layer indexed as ($l-1$). This can be represented as:
\begin{align}
\mathbf{A}_l = \mathcal{T}^{-1}(\mathbf{M} \circ \mathcal{T}(f(\mathbf{F}_{l} \otimes \mathbf{A}_{l-1} + \mathbf{b}_{l})))
\end{align}
Here, $\mathcal{T}$ and $\mathcal{T}^{-1}$ denote the frequency transform and its inverse respectively and $\mathbf{M} \in \mathbb{R}^{h''\times w'' \times m}$ represents the spectral dropout mask. We next describe the transformation and masking operations in detail.

\paragraph{Spectral Masking} The mask $\mathbf{M}$ involved in the spectral dropout comprises of both deterministic and stochastic components as follows:
\begin{align}
& M_{i,j,k} \sim Bernoulli(p') \\
& p' = \left\{
\begin{array}{ll}
      p  &  \forall \{i,j,k\}: T_{i,j,k} > \tau \\
      0 &  \forall \{i,j,k\}: T_{i,j,k} < \tau \\
\end{array} 
\right. \\
& where,\;  T_{i,j,k} \in \mathbf{T} = \mathcal{T}(\mathbf{A}'_l). \notag
\end{align}
$\tau$ is a threshold on the magnitude of the coefficients for frequency dropout. The percentage of activations ($\eta$) above a  fixed threshold $\tau$ can be variable for different input batches. Similarly, the threshold can be adaptive if we want to keep $\eta$ fixed for different inputs. In other words,  either $\tau$ or $\eta$ can be kept unchanged in the following equation:
\begin{align}
\eta = \frac{\sum\limits_{i,j,k} \llbracket T_{i,j,k} > \tau \rrbracket}{h''w''m} .
\end{align}

\paragraph{Spectral Transformation}
 We use a discrete sinusoidal unitary transform with fixed basis functions, known as the DCT-II \cite{ahmed1974discrete}. The main motivation of using DCT is that it is a real-valued transform and very fast algorithms are available for its computation. Furthermore, DCT is an orthogonal and separable frequency domain transform which avoids redundancy by performing signal decorrelation. If $\mathbf{a} = \{a_i, :i \in [1,n]\}$ denotes the CNN activations, we can perform 1D forward and inverse DCT as follows:
\begin{align}
x_k & = \alpha(k) \sum\limits_{i=1}^{n} a_i \beta(i,k), k \in [1,n] \\
a_i & = \sum\limits_{k=1}^{n} \alpha(k) x_k \beta(i,k), i \in [1, n] 
\end{align}
where, $\alpha$ and $\beta$ are defined as:
\begin{align}
\alpha(k) & = \sqrt{1/n} \llbracket k=1 \rrbracket + \sqrt{2/n}\llbracket k\neq 1 \rrbracket \\
\beta(i,k) & = \cos\left[ \frac{\pi(2i-1)(k-1)}{2n}\right].
\end{align}
 Here, $\beta$ denotes the cosine basis functions which are mutually orthogonal. The DCT performs energy compaction and retains most of the signal energy in only a few dominant coefficients.

\begin{figure}
\centering
\hspace{-4mm}\includegraphics[clip=true, trim= 8cm 8.5cm 14cm 3cm, width = 1.05\columnwidth]{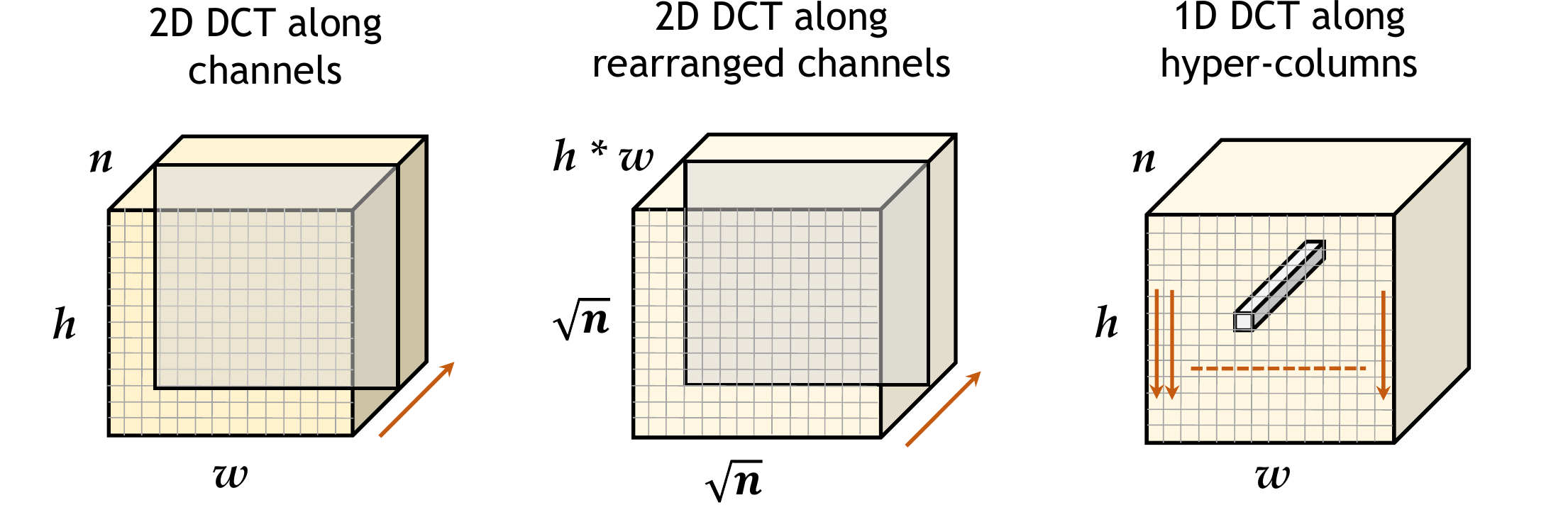}
\caption{Spectral transform variants investigated in this work.}
\label{fig:TransVariants}
\end{figure}

Similarly, for the convolutional layers, a 2D forward and inverse DCT can be defined as:
 \begin{align}
x_{k, \ell} & = \alpha(k)\alpha(\ell) \sum\limits_{i=1}^{n} \sum\limits_{j=1}^{n} a_{i,j}\beta(i,k)\beta(j,\ell), \label{eq:twoD1}\\
a_{i,j} & = \sum\limits_{k=1}^{n}\sum\limits_{\ell=1}^{n} \alpha(k)\alpha(\ell) x_{k,\ell} \beta(i,k)\beta(j,\ell), \label{eq:twoD2}
\end{align}
where, $ \{k,\ell\} \in [1,n]$ and $ \{i,j\} \in [1, n]$ in Eq.~\ref{eq:twoD1} and Eq.~\ref{eq:twoD2}, respectively. 
The 1D and 2D variants are illustrated in Fig.~\ref{fig:TransVariants}.

\paragraph{Frequency Transform as a CNN Layer}
A main bottleneck while introducing frequency domain transformation in the neural network forward and backward signal flow is the computational efficiency. To overcome this, we use a DCT-II transform with fixed basis functions and show that it can be formulated as a convolution operation. This effectively integrates the frequency transformation with-in a regular CNN and requires no modification in the forward and backward signal prorogation mechanisms during the learning and inference process.  

The frequency domain transformation can be applied to the CNN feature maps in three possible ways. \emph{(a)} Given the feature map $\mathbf{A}'_{l}$, a 2D transformation can be applied individually to each channel of the feature map. \emph{(b \& c)} A frequency domain transformation can also be applied to hyper-columns\footnote{We use the term \emph{hypercolumns} to denote  tube-vectors (mode-3 vectors) in the tensor $\mathbf{A}_{l-1}$ of CNN activations.} corresponding to each individual spatial location in the feature map. In this case, either a 2D transform can be applied by rearranging each hyper-column $\mathbf{a}'_{l} \in \mathbb{R}^{m}$ to form $\sqrt{m}\times \sqrt{m}$ dimensional maps ($h'' \times w''$ in total), or only a 1D transform can be applied to each hyper-column by treating it as a 1D signal.

Specifically, the frequency transform $\mathcal{T}$ can be defined as a CNN convolution layer with filter weights $\mathbf{F}_{\mathcal{T}} \in \mathbb{R}^{h'\times w' \times n \times m} $, such that $h'=w'=1$ and we assume a equi-dimensional feature representation in the spectral domain i.e., $n = m$. The filter weights are set as follows:
\begin{align}
\mathbf{F}_{\mathcal{T}}  & = [\mathbf{v}_1,\ldots,\mathbf{v}_m]:\mathbf{v}_i\in \mathbb{R}^m \\
\mathbf{v}_i & = [\alpha(1)\beta(i,1),\ldots, \alpha(m)\beta(i,m)]
\end{align}
 for a 1D DCT-II transform and 
\begin{align}
\mathbf{v}_i & =  vec({\mathbf{v}'_{p}}^T \mathbf{v}'_{q}) : \mathbf{v}'_i \in \mathbb{R}^{\sqrt{m}}, \notag\\
\mathbf{v}'_i & = [\alpha(1)\beta(i,1),\ldots, \alpha(\sqrt{m})\beta(i,\sqrt{m})] \text{, where }  \notag
\end{align}
\begin{align}
p & = \lceil \frac{i}{\sqrt{m}} \rceil, \; 
q'  = i-\sqrt{m}\lfloor \frac{i}{\sqrt{m}}\rfloor,  
\text{and}  \label{eq:p_def}\\
q & = q'\llbracket q'\neq 0\rrbracket + \sqrt{m}\llbracket q'= 0\rrbracket, \label{eq:q_def}
\end{align}
for a 2D DCT-II transform. Similarly for an inverse DCT, we have:
\begin{align}
\mathbf{F}_{\mathcal{T}^{-1}} & = [\hat{\mathbf{v}}_1, \ldots, \hat{\mathbf{v}}_m]: \hat{\mathbf{v}}_m \in \mathbb{R}^{m}, \text{ where } \\
\hat{\mathbf{v}}_i & = [\alpha(i)\beta(1,i),\ldots, \alpha(i)\beta(m,i)]
\end{align}
for a 1D DCT-II and 
\begin{align*}
\hat{\mathbf{v}}_i & =  vec({\hat{\mathbf{v}}{'}_{p}}^T \hat{\mathbf{v}}'_{q}) : \hat{\mathbf{v}}'_i \in \mathbb{R}^{\sqrt{m}},  \\ 
\hat{\mathbf{v}}'_i & = [\alpha(i)\beta(1,i),\ldots, \alpha(i)\beta(\sqrt{m},i)],
\end{align*}
for a 2D DCT-II transform, where $p,q$ are defined similar to Eq.~\ref{eq:p_def} and Eq.~\ref{eq:q_def} respectively.

In our experiments, a DCT transform of the same dimensions as input is applied for simplicity. Note that the only limitation for a 2D transform along feature channels is that the number of channels should satisfy:$\mod(\sqrt{n},1) = 0$.

\section{CNN Architectures}\label{sec:cnn_arch}
We experiment with a number of popular CNN architectures to explore the efficacy of the spectral dropout approach. Rather than pushing state of the art, our main goal is to study the learning behaviors and performance trends of the networks when spectral dropout is applied during the train and test phases. Therefore, we focus on only simpler and standard network architectures proposed in the literature `as is', and plug-in our proposed regularization module to clearly demonstrate the performance improvements.

\begin{table}[!t]
\centering
\caption{CNN Architectures used for evaluation and analysis of spectral dropout approach. In the first column, the first term inside the brackets show number of filters for MNIST while the second term shows number of filters for CIFAR and SVHN.}
\scalebox{1.1}{
\begin{tabular}{@{}c @{\hskip 0.25in} c @{\hskip 0.25in}c @{}l@{}}
\toprule
LeNet & NiN & ResNet \\
\midrule
cnv$-5\times 5$ (20/32)     & cnv$-5\times 5$ (192)   &  cnv$-3 \times 3$ (16)\\
maxpool ($\downarrow$2)  & cnv$-1\times 1$ (160)   &  cnv$-1\times 1$ (16) & \rdelim\}{3}{1mm}[$\times t$]\\
cnv$-5\times 5$ (20/32)     & cnv$-1\times 1$ (96)    &  cnv$-3\times 3$ (16) & \\
maxpool ($\downarrow$2)  & maxpool ($\downarrow$2) &  cnv$-1\times 1$ (64) & \\
cnv$-5\times 5$ (50/64)        & dropout & cnv$-1\times 1$ (32) & \rdelim\}{3}{1mm}[$\times t$]\\
maxpool ($\downarrow$2)    & cnv$-5\times 5$ (192) & cnv$-3\times 3$ (32) & \\
cnv$-4\times 4$ (500/64)   & cnv$-1\times 1$ (192) & cnv$-1\times 1$ (128) & \\
cnv$-1\times 1$ (C)  & cnv$-1\times 1$ (192) & cnv$-1\times 1$ (64) & \rdelim\}{3}{1mm}[$\times t$]\\
softmax loss    & maxpool ($\downarrow$2) & cnv$-3\times 3$ (64) & \\
        & dropout & cnv$-1\times 1$ (256) & \\
  & cnv$-3\times 3$ (192) & avgpool \\
  & cnv$-1\times 1$ (192) &  cnv$-1\times 1$ (C)\\
  & cnv$-1\times 1$ (C) &  softmax loss \\
    & maxpool ($\downarrow$7)  & \\
        & softmax loss & \\
        \bottomrule
\end{tabular}
}
\label{tab:net_arch}
\end{table}

We use three standard CNN architectures namely, \textbf{(a)} LeNet architecture  \textbf{(b)} Network in Network (NiN) and the recent \textbf{(c)} Residual Network (ResNet) with pre-activation units. The architectures of these networks are shown in Table~\ref{tab:net_arch}. The LeNet architecture is slightly different from the one proposed in \cite{lecun1998gradient}, which leads to a lower training error. Each convolution layer in LeNet and NiN is followed by a ReLU layer. For the residual network, each residual unit has a `bottleneck' architecture, as shown inside the braces in Table~\ref{tab:net_arch}. The residual unit is repeated `$t$' times depending on the total weight layers ($L$) in the network, $t = {(L-2)/9}$. For our experiments, we keep $t=18$ ($L=164$). With in each residual block, there exist identity short-cut connections from the input to the output of the plain layers, combined together with a sum layer at the end of each residual block. Each weight layer (except the initial convolution layer) has a pre-activation mechanism consisting of a weight layer preceded by a batch normalization (BN) and a ReLU layer. The pre-activation mechanism leads to a better performance than the original ResNet \cite{he2016identity}.

As a baseline NiN architecture, we used the one proposed in \cite{lin2013network} without two intermediate dropout layers. For experiments involving NiN architecture with Dropout and Drop-Connect regularization, we use these regularizer layers at the same locations as proposed in \cite{lin2013network}. Since other network architectures do not have an ideal predefined location for the regularizer, we place these layers at the same location where spectral dropout has been found to give an optimal performance on a validation set. Note that the spectral dropout is plugged in different architectures which sometimes already have a regularization mechanism e.g., Dropout in NiN and Batch Normalization in ResNet. Based on the improvements described in Sec.~\ref{sec:exp_analysis}, the spectral dropout contributes to better performance both with and without other regularizers  (e.g., in LeNet and NiN respectively).

\section{Experimental Analysis}\label{sec:exp_analysis}
\subsection{Overview}
In all our experiments, we report performances of single networks (LeNet, NiN and ResNet). Note that previous literature reports on the use of different augmentation arrangements for different datasets to obtain an optimal performance. Our goal here in not to establish new state of the art, but to make a fair comparison with other related approaches and demonstrate the performance gain with spectral dropout. Therefore, we do not use any form of data augmentation in our experiments. In principle, the use of data augmentation with spectral dropout should result in better performance. 

In Sec.~\ref{sec:SpectralDropout}, we described two possible ways to perform spectral masking. An almost comparable performance was noted by keeping either of the two variables ($\tau$ and $\eta$) fixed. However, an adaptive $\tau$ required more computational resources. We therefore keep it fixed during our experiments.  
For each dataset, we used a held-out validation set ($20\%$ of the training set) to tune the position of the spectral dropout block, the threshold parameter $\tau$ and the learning rates. Similar to \cite{lin2013network}, we retrain the networks from scratch on the complete training data once these parameters are fixed. Other hyper-parameters were kept same during all experiments e.g., batch size ($200$), number of epochs ($200$) and weight decay ($5\times10^{-4}$). A mean image was subtracted from the input images in all cases.

\subsection{MNIST} This dataset contains $70,000$ gray-scale images of handwritten digits ($0-9$) with size $28\times 28$. There are $60,000$ images for training and another $10,000$ for testing.  For consistency of architectures used in our experiments, we resized $28\times 28$ MNIST images to $32\times 32$ to match with the image size available in other datasets.

The LeNet architecture for MNIST is slightly different compared to the one used for CIFAR and SVHN datasets (see Table~\ref{tab:net_arch}). More precisely, the number of convolutional filters in the initial three weight layers is slightly less than the number used for experiments on CIFAR and SVHN datasets with LeNet architecture.  This is because the MNIST being a relatively smaller dataset needs less parameters in the initial weight layers.  

We report our results and comparisons in Table~\ref{tab:MNIST_perf}. For each baseline architecture, we report performances with Dropout, Drop-Connect, our proposed spectral dropout, and without any of these regularization modules (baseline).  We achieve a significant reduction in error on MNIST dataset using the spectral dropout technique, which sets a new state of the art for single model performance using LeNet, NiN and ResNet architectures without any data augmentation. 

We also experiment with different variants of spectral transformation on the MNIST dataset (see Table~\ref{tab:DCT_variants}). We note that although there is minimal difference in performance for different DCT variants, however, 2D DCT along hyper-columns achieve lowest error rate of $0.51\%$ with the LeNet model. Therefore, in our experiments on CIFAR-10 and SVHN datasets, we use a 2D DCT transformation along the output activation hyper-columns. 

\begin{table}
\centering
\caption{Results on MNIST Dataset: All performances are reported for a single network with no data augmentation.}
\scalebox{1.1}{
\begin{tabular}{l c c c}
\toprule
\multicolumn{1}{l}{Method} & \multicolumn{3}{c}{Error (\%)} \\
\midrule
\multicolumn{1}{l}{Drop-Connect \cite{wan2013regularization}} & \multicolumn{3}{c}{0.63} \\
\multicolumn{1}{l}{Stochastic Pooling \cite{zeiler2013stochastic}} & \multicolumn{3}{c}{0.47} \\
\multicolumn{1}{l}{Maxout Networks \cite{goodfellow2013maxout}} & \multicolumn{3}{c}{0.45} \\
\multicolumn{1}{l}{Deeply-Supervised Net \cite{lee2015deeply}} & \multicolumn{3}{c}{0.39} \\
\midrule
Model ($\rightarrow$)         & LeNet & NiN & ResNet \\
\midrule
Baseline & 0.93 & 0.64 & 0.55\\
with Dropout  &  0.71 &  0.45 & 0.44\\
with Drop-Connect  & 0.77 & 0.46 &  0.42\\
with Spectral Dropout  &  \cellcolor{gray!25}{0.51} & \cellcolor{gray!25}{0.40} &  \cellcolor{gray!25}{0.38} \\
\bottomrule
\end{tabular}}
\label{tab:MNIST_perf}
\end{table}

\begin{table}
\centering
\caption{Comparison of results with different variants of spectral transformation on the MNIST digits dataset.}
\scalebox{1.1}{
\begin{tabular}{l c c c}
\toprule
Variant & 1D-DCT(H) & 2D-DCT(H) & 2D-DCT(C) \\
\midrule
Error (\%) & 0.52 & \cellcolor{gray!25} 0.51 & 0.55\\
\bottomrule
\end{tabular}}
\label{tab:DCT_variants}
\end{table}

\subsection{CIFAR-10} The CIFAR-10 dataset contains $60,000$ color images of $32\times 32$ for ten object classes. There are $50,000$ images for training and another $10,000$ for testing.  

The results and comparisons are summarized in Table~\ref{tab:CIFAR10_perf}. Data whitening and contrast normalization were applied as a preprocessing step. Our approach achieves a consistent boost in performance for all the three network architectures. We note that random dropout was mostly outperformed by the Drop-Connect approach by a small margin on the CIFAR-10 dataset. We do not perform hyperparameter tuning for local receptive field size and weight decay. This is the reason why we achieved slightly lower performance numbers for NiN compared to those reported in \cite{lin2013network}.

\subsection{SVHN}
We experiment on the Street View House Numbers (SVHN) dataset which contains $604,388$ training and $26,032$ testing images. The images are centered on the digits $0-9$ (ten classes) in the MNIST-like $32\times 32$ format. But different from MNIST, the dataset contains color images with size and appearance variations along-with presence of distortions especially near the image periphery.  

Similar to previous works \cite{lin2013network,zeiler2013stochastic}, we perform contrast normalization as a preprocessing step. Note that we obtain a slightly larger error rate on SVHN using NiN architecture because the validation set was used only to tune the learning rates (keeping other hyper parameters same as before). Still, our approach consistently performed better than Dropout and Drop-Connect.

\begin{table}
\centering
\caption{Results on the CIFAR-10 Dataset (using a single network and no data augmentation).}
\scalebox{1.1}{
\begin{tabular}{l c c c}
\toprule
\multicolumn{1}{l}{Method} & \multicolumn{3}{c}{Error (\%)} \\
\midrule
\multicolumn{1}{l}{Stochastic Pooling \cite{zeiler2013stochastic}} & \multicolumn{3}{c}{15.1} \\
\multicolumn{1}{l}{Maxout Networks \cite{goodfellow2013maxout}} & \multicolumn{3}{c}{11.7} \\
\multicolumn{1}{l}{Drop-Connect (with aug) \cite{wan2013regularization}} & \multicolumn{3}{c}{11.1} \\
\multicolumn{1}{l}{Deeply-Supervised Net \cite{lee2015deeply}} & \multicolumn{3}{c}{9.69} \\
\midrule
Model ($\rightarrow$)    & LeNet & NiN & ResNet \\
\midrule
Baseline & 19.2 & 15.8 & 6.02 \\
with Dropout  & 17.3 & 11.4 & 5.97 \\
with Drop-Connect  &  17.0 & 10.9 & 5.99 \\
with Spectral Dropout  &  \cellcolor{gray!25}{16.3} & \cellcolor{gray!25}{9.14} & \cellcolor{gray!25}{5.76}\\
\bottomrule
\end{tabular}}
\label{tab:CIFAR10_perf}
\end{table}
\begin{table}
\centering
\caption{Results on the SVHN Dataset (using a single network and no data augmentation).}
\scalebox{1.1}{
\begin{tabular}{l c c c}
\toprule
\multicolumn{1}{l}{Method} & \multicolumn{3}{c}{Error (\%)} \\
\midrule
\multicolumn{1}{l}{Stochastic Pooling \cite{zeiler2013stochastic}} & \multicolumn{3}{c}{2.80} \\
\multicolumn{1}{l}{Maxout Networks \cite{goodfellow2013maxout}} & \multicolumn{3}{c}{2.47} \\
\multicolumn{1}{l}{Drop-Connect \cite{wan2013regularization}} & \multicolumn{3}{c}{2.23} \\
\multicolumn{1}{l}{Deeply-Supervised Net \cite{lee2015deeply}} & \multicolumn{3}{c}{1.92} \\
\midrule
Model ($\rightarrow$)         & LeNet & NiN & ResNet \\
\midrule
Baseline & 4.29 & 2.74 & 2.22\\
with Dropout  &  4.17 & 2.72 &  2.17\\
with Drop-Connect  & 4.18 &   2.72 & 2.19\\
with Spectral Dropout  & \cellcolor{gray!25}{4.05} & \cellcolor{gray!25}{2.67} &  \cellcolor{gray!25}{2.12}\\
\bottomrule
\end{tabular}}
\label{tab:SVHN_perf}
\end{table}

\subsection{Test-time Spectral Dropout}
At test time, the deep CNN model only gives point estimates without any uncertainty information.  This lack of information about uncertainty makes it difficult to estimate the confidence level of a prediction. Therefore, similar to recent works which use dropout as a tool to obtain uncertainty estimates \cite{kendall2015bayesian,Gal2015DropoutB}, we apply multiple runs of spectral dropout at test time to study the resulting uncertainty estimates and performance gains. Since spectral dropout incorporates stochastic dropout in the spectral domain,  the uncertainty bounds are approximation  of the estimates from a Gaussian process \cite{Gal2015DropoutB}.

\begin{figure}\centering
\includegraphics[trim=4.8cm 4.5cm 3.5cm 2.5cm, clip=true, width=0.9\columnwidth]{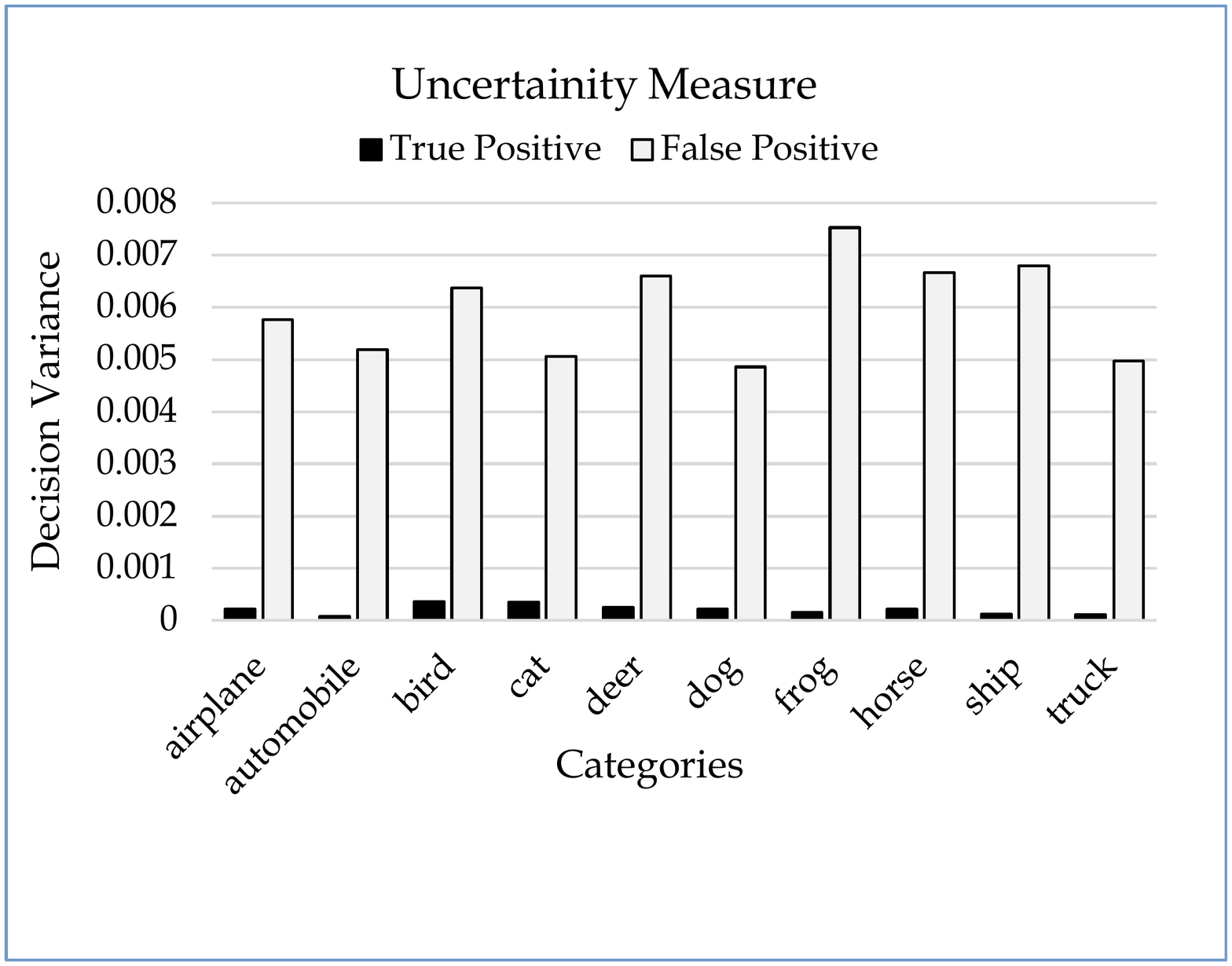}
\caption{Plot of uncertainty measures for all categories in the CIFAR-10 dataset.}
\label{fig:tstSD}
\end{figure}

Let us suppose that we obtain a set of predictions by running $T$ iterations of spectral dropout network with the same input $\mathbf{x}_i$ to obtain $ \{\hat{\mathbf{y}}_i^{t}\} : t\in[1,T]$.  The moments of the empirical posterior distribution are useful for our purpose. 
The final class decisions are established using predictive mean (Eq.~\ref{eq:pred_mean}) and the uncertainty is accounted by variance (Eq.~\ref{eq:pred_var}), as follows:
\begin{align}
\mathbf{y}_i & = \mathbb{E}[\hat{\mathbf{y}}_i] = \frac{1}{T}\sum_{t=1}{\hat{\mathbf{y}}_i^t} \label{eq:pred_mean} 
\end{align}
\begin{align}
var(\hat{\mathbf{y}}_i) & = \frac{1}{T}\sum_{t=1}^{T}{\left(\hat{\mathbf{y}}_i^t\right)^T \hat{\mathbf{y}}_i^t} - \mathbb{E}[\hat{\mathbf{y}}_i]^T \mathbb{E}[\hat{\mathbf{y}}_i] \label{eq:pred_var}
\end{align}
Figure~\ref{fig:tstSD} shows the averaged uncertainty measures of both correct and incorrect predictions for all classes in the CIFAR-10 dataset. The correct predictions are highly confident compared to the incorrect predictions, which are largely uncertain (10$\times$ to 40$\times$). Table~\ref{tab:tstSD} shows the performance improvement with the increasing number of spectral dropout runs for each sample. We found that although the uncertainty estimates are reliable, we get slightly better performance using the  majority voting: 
$ mode( \{\hat{z}_i^1 \ldots \hat{z}_i^T \}) $ compared with of ${z}_i = \underset{c}{\operatorname{argmax}}\;  y_i^c$.

\begin{table}
\centering
\caption{Effect of test-time spectral dropout on the error rate for CIFAR-10 dataset using NiN architecture. }
\scalebox{1.1}{
\begin{tabular}{l c c c c c}
\toprule
Runs & $10^0$ & $10^1$ & $10^2$ & $10^3$ & $10^4$ \\
\midrule
Error (\%) & 9.14 & 9.11 & 9.06 & 9.04 & \cellcolor{gray!25} 9.03 \\
\bottomrule
\end{tabular}}
\label{tab:tstSD}
\end{table}

\section{Analysis and Discussion} \label{sec:analysis}

\subsection{Positional Analysis}\label{sec:Positional Analysis}

\begin{figure}\centering
\includegraphics[trim=9cm 2cm 8cm 2cm, clip=true, width=0.9\columnwidth]{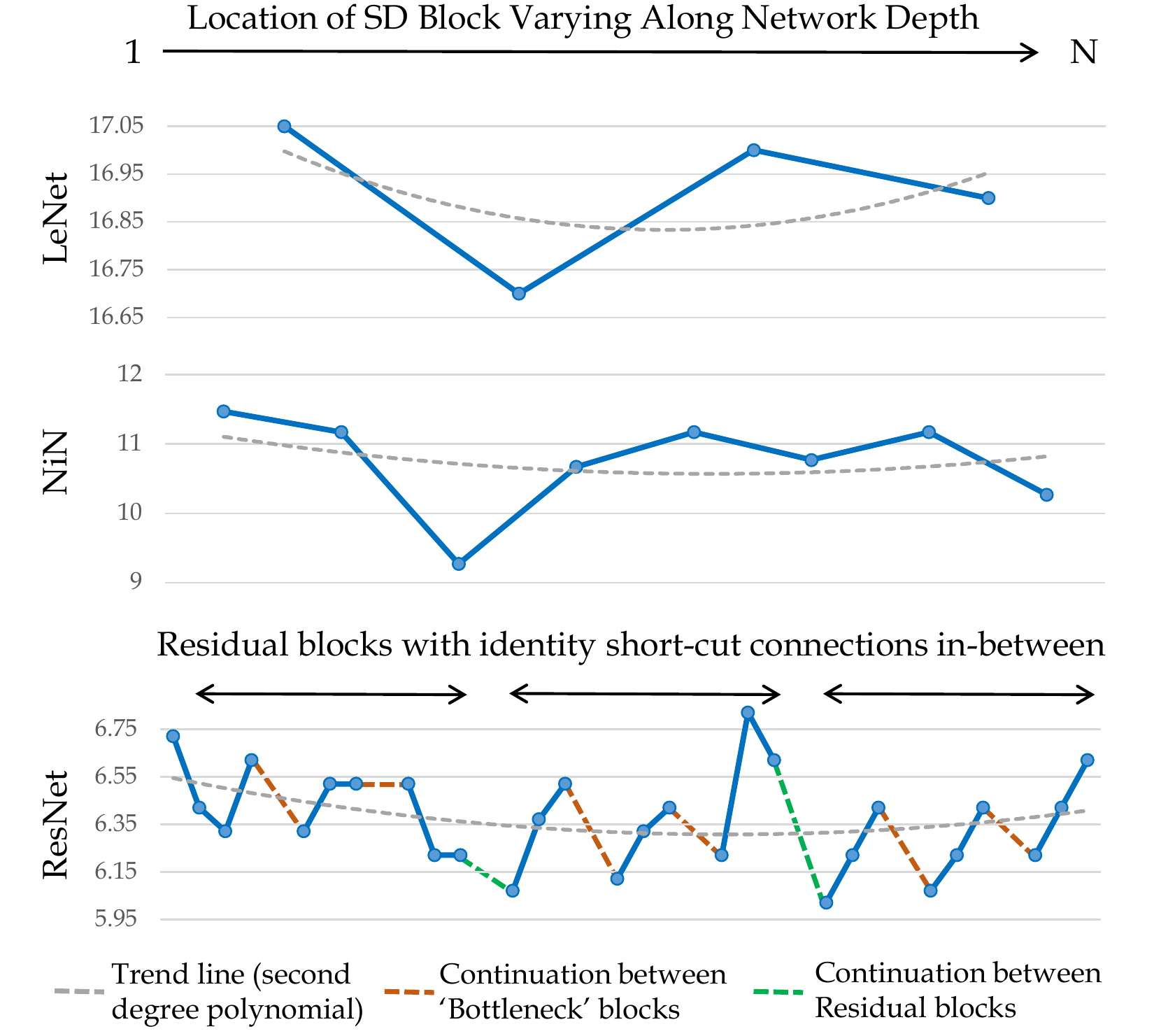}
\caption{Performance trend with different locations of SD layer on the CIFAR-10 dataset.}
\label{fig:position}
\end{figure}

The effect of the position of spectral dropout block within the deep network has been analyzed in Fig.~\ref{fig:position}. For each of the three CNN architectures, namely LeNet, NiN and ResNet, we report the performance trend on the validation set versus the position of spectral dropout  module. We note that the best results are obtained when the spectral dropout is applied at the intermediate levels of abstraction. To emphasize this trend, we fit a second degree polynomial on the achieved performances. This is in contrast to the regular Dropout approach, which is usually applied at the end of a deep network (adjacent to fully connected layers). 
Specifically, for the case of ResNet, we observed a  decrease in performance when the position of spectral dropout  module was shifted from the first layer in the bottleneck block towards the last layer. Note that for each residual block, we analyzed error rate by plugging in the spectral dropout  module after first, middle and the last bottleneck blocks (separated by dotted line). The best performance was observed when the spectral dropout was applied after the first convolution layer in the last residual block. Note that all the  module positions in Fig.~\ref{fig:position} are on the highway signal path. We also experimented with spectral dropout  block in the short-cut connections within a residual network. However, this resulted in a decrease in the overall performance. This observation is consistent with the literature \cite{he2016identity}, where it has been found that modifying the identity connections results in performance degradation. 

Furthermore, we also experimented with \emph{multiple} spectral dropout blocks within each of the LeNet, NiN and ResNet architecture, which did not provide much performance gain while increasing the network convergence time and computational requirements.

\subsection{Computational Complexity}
The spectral dropout block comprises of two convolutional layers, which adds computational overhead of $\mathcal{O}(2h''w''nm)$. However, it is important to note that vector multiplication is implemented as an efficient block matrix multiplication routine in BLAS library, which is faster than directly computing FFT and IFFT on a CPU. On the CIFAR-10 dataset with a batch size of 200, using a core i7 machine with a NVIDIA Titan-X 12GB card and 16GB RAM, it takes 127.9$\mu$s and 44.7$\mu$s respectively to process one image with a spectral dropout CNN during the train and test phases.  In comparison, a baseline CNN takes 110.3$\mu$s and 41.0$\mu$s, while a dropout CNN takes 125.9$\mu$s and 41.2$\mu$s during the train and test phase, respectively.


\subsection{Sparsity Analysis}
\begin{figure}\centering
\includegraphics[trim = 0.5cm 9.5cm 20cm 0.5cm, clip=true, width=0.9\columnwidth]{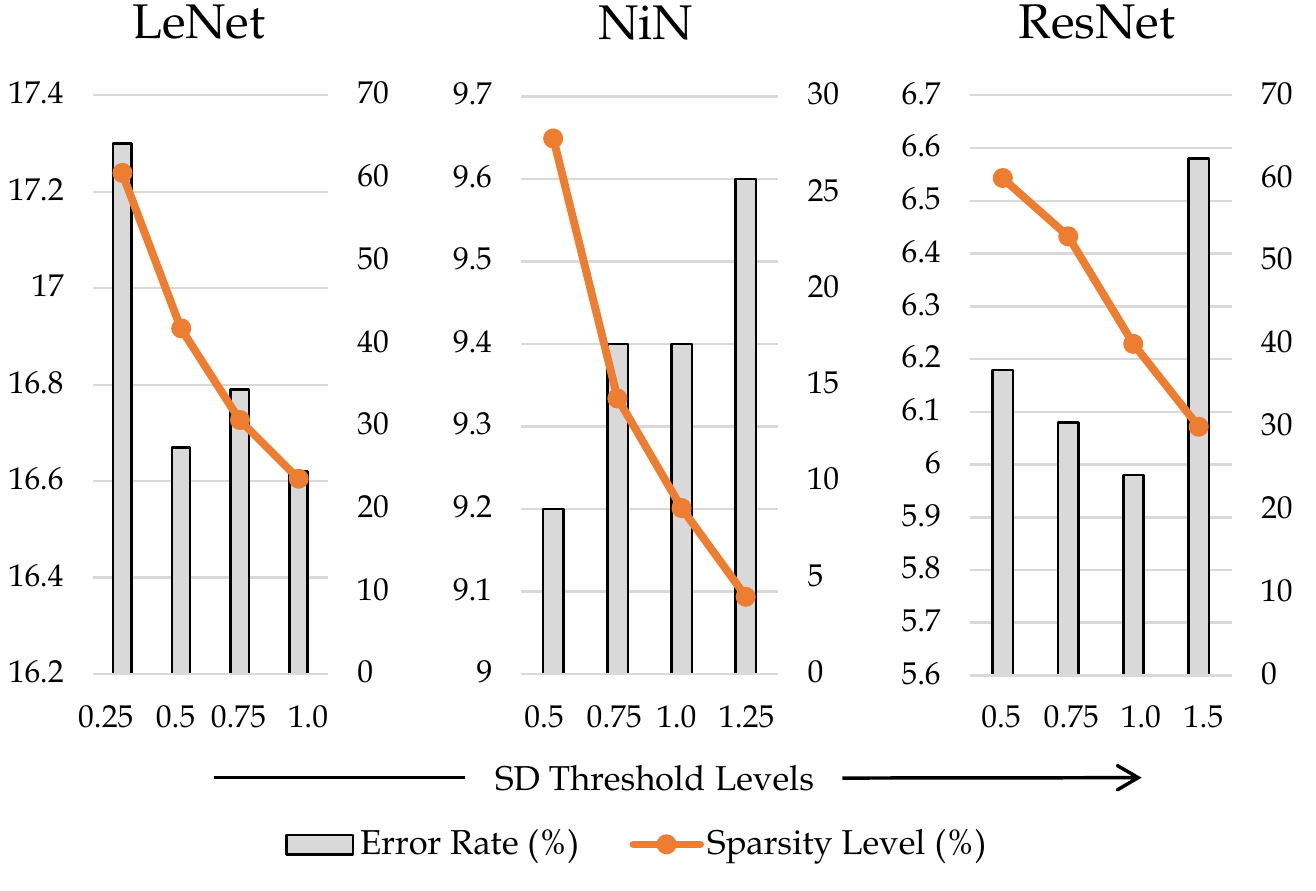}
\caption{Performance and sparsity trend for different thresholds of SD on the CIFAR-10 dataset. The axis scale on left denotes error rate while the right scale represents the percentage of retained neural activations.}
\label{fig:trendSDth}
\end{figure}

We analyze the impact of spectral dropout threshold on the percentage of pruned neural activations and the corresponding performance levels. Figure~\ref{fig:trendSDth} summarizes the trend on CIFAR-10 dataset with LeNet, NiN and ResNet architectures. The analysis is performed on a validation set similar to the positional analysis. The comparison is shown for the best SD position in each network as identified in Sec.~\ref{sec:Positional Analysis}. We note that while the performance is sensitive to the percentage of pruned activations, there exists a consistency in the activation levels for different inputs and network architectures. For the LeNet and NiN architectures, the best performance was achieved when $\sim 70\%$ of the activations were pruned. For the ResNet architecture, we noted the best performance at a slightly lower pruning rate i.e., $\sim 60\%$. Due to this consistency, we found that setting a fixed activation threshold level performs identical to a threshold on the sparsity level, but with considerable gains in computational efficiency. It is also interesting to note that we apply random dropout with a relatively low pruning probability ($20\%$) on top of the threshold based pruning on spectral activations. Therefore, with much high sparsity levels ($\sim 70-80\%$) compared to random dropout (which normally deactivates $50\%$ of the neurons), we are able to achieve lower error rates with much faster convergence during the network training.

 \subsection{Effect on Convergence Time}
\begin{figure}
\centering
\includegraphics[width=0.48\columnwidth]{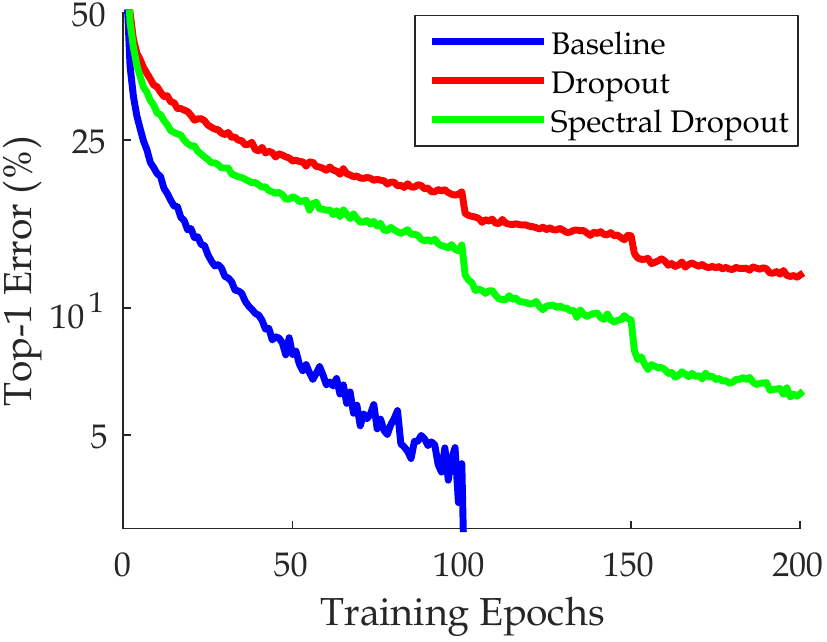}
\includegraphics[width=0.48\columnwidth]{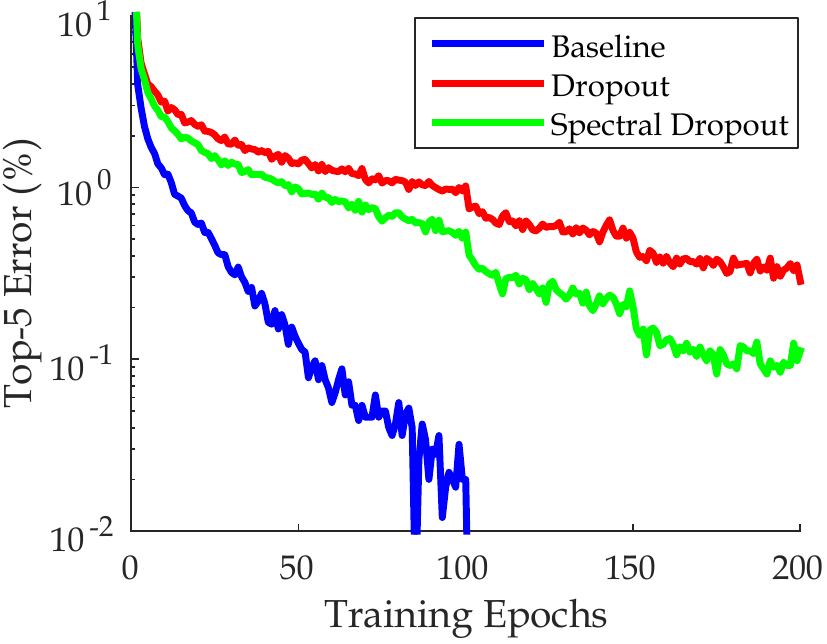}
\caption{Network convergence rate during training.}
\label{fig:conv_time}
\end{figure}

The spectral dropout speeds up the convergence rate during the training process. In Fig.~\ref{fig:conv_time}, top-1 and top-5 error rates for the CIFAR-10 dataset are shown. Note that the baseline model converges rapidly, but does not generalize well to the unseen data. Compared to the Dropout, spectral dropout not only achieves better performance but also enhances the convergence speed. Similar convergence behavior can be seen for the MNIST dataset in Fig.~\ref{fig:mnist_err}.

\subsection{Power Analysis}
\begin{figure}
\centering
\includegraphics[trim=3.5cm 2.7cm 3.5cm 2.5cm, clip=true, width=0.85\columnwidth]{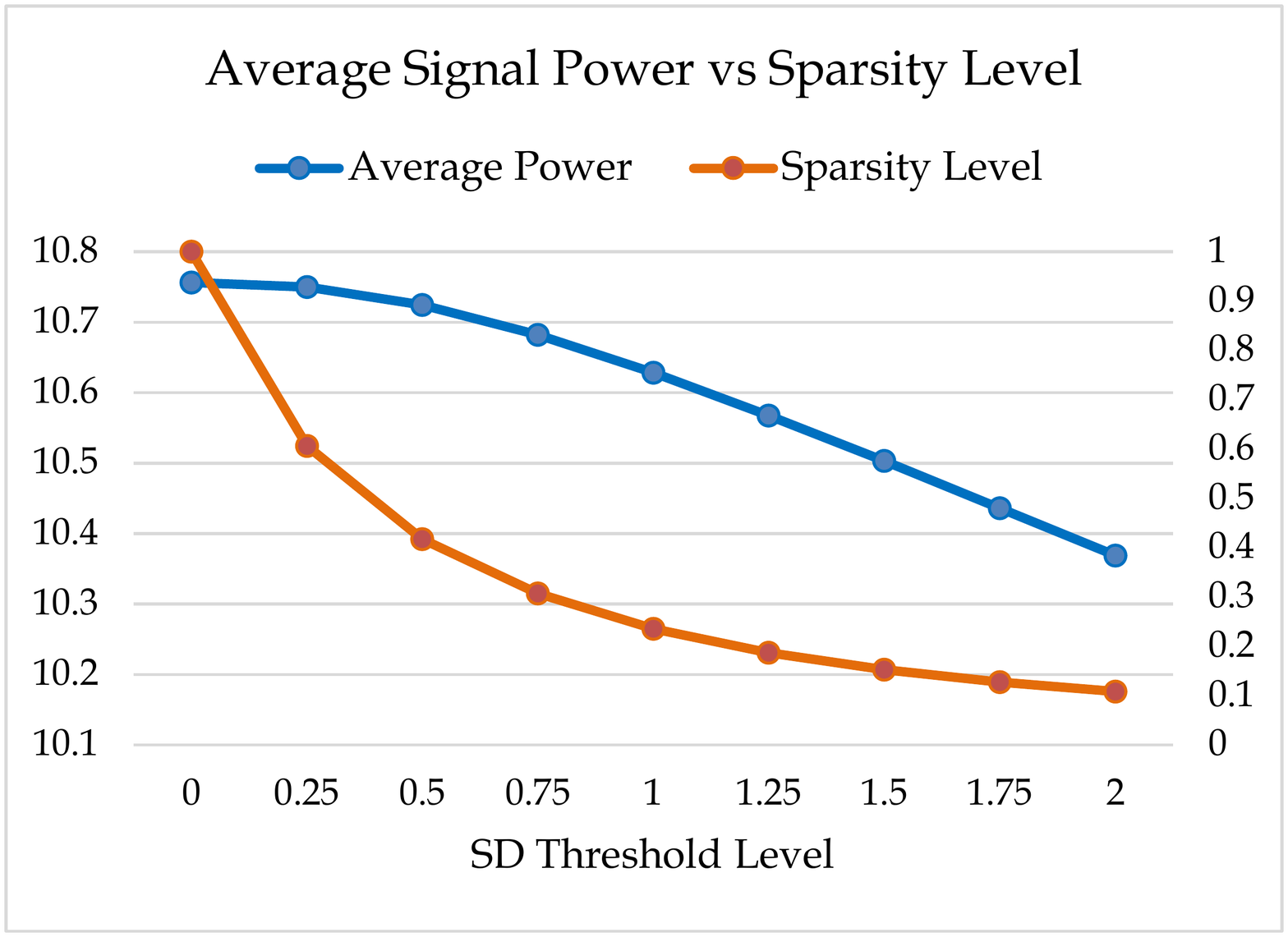}
\caption{Analyzing the effect of sparsity level on the signal power. The axis scale on left shows average power, while the scale on right indicates percentage of retained neural activations. }
\label{fig:power}
\end{figure}

We study the effect of different spectral dropout thresholds on the retained power of the feature vectors propagating through the network (see Fig.~\ref{fig:power}). Since the signal power and its energy are directly proportional, this analysis holds true for energy as well. With the increase in the percentage of pruned neurons, we notice a consistent decline in signal power. However, this decline is not significant when compared to original power, i.e., only a $3.6\%$ drop in signal power as a result of $90\%$ pruning. This explains why the network convergence rate is relatively higher compared to the Dropout approach. 

\section{Conclusion}
We introduced a new approach to perform regularization in the deep neural networks in an efficient manner.  The proposed approach uses spectral domain transformation to identify and ignore weak frequency coefficients such that the co-adaptations of the feature detectors are avoided. Furthermore, we show that the spectral domain transformation can be formulated as a convolution operation, thus enabling computational efficiency and an end-to-end trainable network. Our results demonstrate a superior performance compared to other regularization methods and baseline approaches on a range of popular CNN architectures and image classification datasets.

%

%
%

\ifCLASSOPTIONcaptionsoff
  \newpage
\fi

\bibliographystyle{IEEEtran}
\bibliography{egbib}

%
%
%
%
%
%

\end{document}